\documentclass[conference]{IEEEtran}
\IEEEoverridecommandlockouts
\usepackage{cite}
\usepackage{amsmath,amssymb,amsfonts}
\usepackage{algpseudocode} 
\usepackage{graphicx}
\usepackage{textcomp}
\usepackage{xcolor}
\usepackage{import}
\usepackage{url}
\usepackage{multirow}
\usepackage{algorithm}
\usepackage{textgreek}
\usepackage[compatibility=false]{caption}
\usepackage{cuted}
\usepackage{hyperref}
\usepackage{booktabs}
\usepackage[table]{xcolor}  
\usepackage{makecell}

\def\BibTeX{{\rm B\kern-.05em{\sc i\kern-.025em b}\kern-.08em
    T\kern-.1667em\lower.7ex\hbox{E}\kern-.125emX}}

\begin{document}

\title{Enhancing Heavy Rain Nowcasting with Multimodal Data: Integrating Radar and Satellite Observations}

% \author{\IEEEauthorblockN{Rama Kassoumeh}
% \IEEEauthorblockA{\textit{Bochum Institute of Technology} \\
% rama.kassoumeh@bo-i-t.de}
% \and
% \IEEEauthorblockN{David Rügamer}
% \IEEEauthorblockA{\textit{LMU Munich} \\
% \IEEEauthorblockA{\textit{Munich Center for Machine Learning}}
% david.ruegamer@stat.uni-muenchen.de}
% \and
% \IEEEauthorblockN{Henning Oppel}
% \IEEEauthorblockA{\textit{Okeanos Smart Data Solutions GmbH} \\
% henning.oppel@okeanos.ai}
% }

\newcommand{\namesep}{\hspace{0.8em}}

\author{
Rama Kassoumeh $^{1}$\namesep
David Rügamer $^{2,3}$\namesep
Henning Oppel $^{4}$\namesep % <- this stop the space

\vspace{0.3em} \\

{\normalsize
  $^{1}$Bochum Institute of Technology\namesep
  $^{2}$LMU Munich\namesep
  $^{3}$Munich Center for Machine Learning\namesep
  $^{4}$Okeanos Smart Data Solutions GmbH
  }
\vspace{0.3em} \\

{\normalsize
  rama.kassoumeh@bo-i-t.de\namesep
  david.ruegamer@stat.uni-muenchen.de\namesep
  henning.oppel@okeanos.ai
  }
}%

\maketitle

\begin{abstract}
The increasing frequency of heavy rainfall events, which are a major cause of urban flooding, underscores the urgent need for accurate precipitation forecasting—particularly in urban areas where localized events often go undetected by ground-based sensors. In Germany, only 17.3\% of hourly heavy rain events between 2001 and 2018 were recorded by rain gauges, highlighting the limitations of traditional monitoring systems. Radar data are another source that effectively tracks ongoing precipitation; however, forecasting the development of heavy rain using radar alone remains challenging due to the brief and unpredictable nature of such events. Our focus is on evaluating the effectiveness of fusing satellite and radar data for nowcasting. We develop a multimodal nowcasting model that combines both radar and satellite imagery for predicting precipitation at lead times of 5, 15, and 30 minutes. We demonstrate that this multimodal strategy significantly outperforms radar-only approaches. Experimental results show that integrating satellite data improves prediction accuracy, particularly for intense precipitation. The proposed model increases the Critical Success Index (CSI) for heavy rain by 4\% and for violent rain by 3\% at a 5-minute lead time. Moreover, it maintains higher predictive skill at longer lead times, where radar-only performance declines. A qualitative analysis of the severe flooding event in the state of North Rhine-Westphalia, Germany on July 14, 2021 further illustrates the superior performance of the multimodal model. Unlike the radar-only model, which captures general precipitation patterns and light rainfall, the multimodal model yields more detailed and accurate forecasts for regions affected by heavy rain. This improved precision is critical for enabling timely, life-saving, and reliable early warnings during heavy rain events that have been repeatedly hitting Europe. The model implementation is available on GitHub
\footnote{\url{https://github.com/RamaKassoumeh/Multimodal_heavy_rain}}.
\end{abstract}


% Include chapters
\section{Introduction}

\begin{figure*}[btp]
    \centering
    \begin{minipage}{0.3\textwidth}
        \centering
        \includegraphics[width=\textwidth]{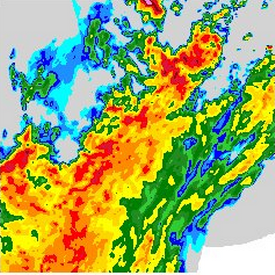}
        \\Ground Truth
    \end{minipage}
    \hfill
    \begin{minipage}{0.3\textwidth}
        \centering
        \includegraphics[width=\textwidth]{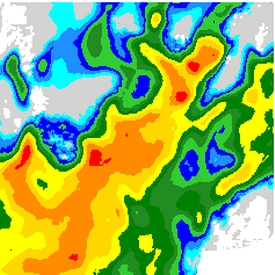}
        \\Radar Model Prediction
    \end{minipage}
    \hfill
    \begin{minipage}{0.3\textwidth}
        \centering
        \includegraphics[width=\textwidth]{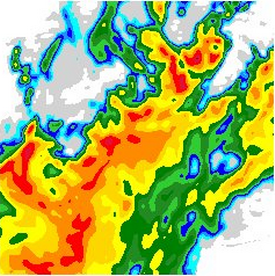}
        \\Multimodal Model Prediction
    \end{minipage}
    \caption{The figure shows the ground truth of a flooding event in the state of North Rhine-Westphalia, Germany, on July 14, 2021, along with predictions from two models: Radar-only and Multimodal (Radar+Satellite) at a 30-minute lead time. Orange and red areas indicate heavy rainfall, with red representing higher precipitation intensities. This event led to catastrophic flooding in the region. The Multimodal model outperforms the Radar-only model, particularly in regions of intense rainfall where the Radar-only model fails to capture the full magnitude of the event. These results demonstrate that satellite data is crucial for accurate prediction of heavy rain scenarios. Capturing the correct magnitude of such events is essential for issuing timely warnings and enabling an effective first response, especially in light of similar extreme weather events that have recently occurred across Europe.}
    \label{fig:compare_30_min}
\end{figure*}

The prediction of heavy rainfall has become increasingly important in meteorological and hydrological research, especially in light of the rising frequency and intensity of such events due to climate change \cite{kotz2024constraining}. Accurate precipitation measurements are essential for real-time flood warning systems, yet current approaches often face limitations. Radar systems, while providing broader spatial coverage than ground-based sensors, measure precipitation indirectly at altitude. This introduces significant uncertainty at ground level, often leading to the underestimation of localized heavy rain events \cite{bpb2021jahrhunderthochwasser}.

Radar technology is a cornerstone of modern nowcasting systems due to its high spatial and temporal resolution—typically 1–2 km² every 5–10 minutes—and its ability to scan large areas \cite{radar}. This makes it well-suited for monitoring fast-evolving weather phenomena and issuing early warnings for convective events. Moreover, radar can detect atmospheric disturbances like thunderstorms before they produce measurable surface rainfall, providing valuable lead time for flood prevention \cite{radar}. However, radar also has limitations. It struggles to pinpoint the timing and location of initial raindrop formation, especially under complex summer conditions influenced by variables such as cloud height, instability, temperature, and humidity \cite{canovas2018assessment}.

Even traditional ground-based rain gauges face critical limitations. Their measurements are restricted to a very small area—just a few meters around the sensor—and cannot capture the spatial heterogeneity of rainfall. For instance, in Germany, gauges detected only 17.3\% of hourly heavy rainfall events between 2001 and 2018 \cite{radar}. This sparse coverage increases the risk of undetected extreme weather, particularly in urban regions with insufficient drainage infrastructure. The catastrophic floods in Germany in July 2021, which caused over 180 fatalities \cite{bpb2021jahrhunderthochwasser}, and the floods in Austria in September 2024, where damages exceeded €1.3 billion and affected over 800 companies \cite{Austria2024}, emphasize the urgent need for more precise and timely heavy rainfall forecasting.

To better capture these fast-evolving weather conditions (especially during summer when heat and moisture trigger sudden downpours), meteorologists are increasingly turning to data fusion approaches. By combining radar and satellite data, forecasts can become more robust and spatially comprehensive.

Satellite imagery provides valuable complementary information by capturing a wide range of atmospheric and surface variables, such as land-water boundaries, snow and ice coverage, soil moisture, and vertical profiles of temperature and humidity \cite{eumetsat}. Different spectral bands (visible, infrared, thermal, microwave) each contribute unique insights: infrared channels help estimate cloud-top temperatures and thickness, while microwave bands offer visibility into water vapor content and precipitation profiles. These capabilities make satellite data an essential component in both qualitative analysis and quantitative prediction of heavy rainfall events \cite{eumetsat}.

This paper focuses on processing satellite data from EUMETSAT and radar data from national sources, and on developing models that combine both modalities. The main objective is to evaluate whether integrating satellite and radar data can improve the accuracy of short-term rainfall predictions. To ensure a fair comparison, we train two models based on a fixed architecture: one using only radar data, and another multimodal model using both radar and satellite data to assess the contribution of satellite information. The effectiveness of this multimodal approach is illustrated in Fig.~\ref{fig:compare_30_min}, which compares predictions from both models during the catastrophic flooding event in North Rhine-Westphalia, Germany, on July 14, 2021.

\section{Related Work}

This section reviews notable approaches for short-term precipitation nowcasting, with a focus on deep learning architectures designed to model spatiotemporal dynamics in weather data. The task is commonly framed as a spatiotemporal sequence forecasting problem, where the goal is to capture both temporal evolution and spatial context.

Recent research in precipitation nowcasting has explored various input–output modalities to improve forecast accuracy. Traditional models often perform radar-to-radar prediction, using sequences of past radar reflectivity frames to forecast future ones and capture local precipitation dynamics directly. Other approaches focus on satellite-to-satellite forecasting, predicting future satellite imagery based on prior observations to model the temporal evolution of cloud and atmospheric features. Satellite-to-radar models have also emerged, aiming to infer radar reflectivity fields directly from satellite observations, enabling precipitation forecasts in areas with limited radar coverage. Multimodal models combining both satellite and radar inputs to predict future radar frames leverage the broad spatial coverage and spectral richness of satellites together with the direct precipitation measurements from radar, enhancing both spatial generalization and event-specific accuracy.

ConvLSTM \cite{convlstm} and RainNet \cite{rainnet} are radar-to-radar models that forecast future radar reflectivity maps from past observations. ConvLSTM extends traditional LSTMs by integrating convolutional operations into its gating mechanisms, enabling the joint modeling of spatial and temporal dependencies. RainNet, on the other hand, adopts a U-Net-like encoder–decoder architecture with skip connections to preserve spatial details and uses a log-cosh loss function to better capture extreme precipitation values. Both models are well-suited for spatiotemporal precipitation nowcasting and have influenced many follow-up works.

As a satellite-to-satellite forecasting model, DeePSat \cite{DeePS} predicts future satellite imagery using sequences of EUMETSAT composites. It processes multi-channel inputs, including Day Microphysics, Severe Storm, and Water Vapor bands, to capture the temporal evolution of cloud and atmospheric structures. The model’s performance has been assessed using Mean Absolute Error (MAE) and Normalized MAE (NMAE), demonstrating its effectiveness in short-term satellite image prediction for nowcasting applications.

WeatherFusionNet \cite{weatherfusionnet} is a satellite-to-radar model that predicts future radar frames using sequences of satellite images. While the architecture includes a module designed to take combined radar and satellite inputs, the final model uses only satellite data due to the poor quality of available radar observations. It leverages a Sat2Rad U-Net to extract precipitation-relevant features, followed by PhyDNet to model temporal dynamics, and a final U-Net to generate the radar-based forecast. A similar two-stage idea is explored in Sat2Rdr \cite{park2025data}, where the model first predicts future satellite frames from past satellite sequences and then uses a separate satellite-to-radar model to convert the predicted satellite frames into precipitation radar images.

Our approach focuses on multimodal nowcasting, combining both satellite and radar inputs to predict future radar frames. Unlike methods that rely solely on satellite data, we utilize reliable radar observations to guide more accurate predictions. We conduct a comparative analysis of radar-only and radar-satellite models to assess the benefit of incorporating satellite imagery. While other studies often aim for year-round precipitation forecasting, our work specifically targets heavy rainfall during the summer period, as these events are among the most difficult to predict and are critical for the effectiveness of early warning systems.

\section{Methodology}
This section describes the data sources, the preprocessing, normalization procedures and evaluation metrics. It also presents the model architectures and loss functions employed in this study.

\subsection{Data sources}

This study focuses on the North Rhine-Westphalia (NRW) region in Germany during the summer months (May to August), when high temperatures increase evaporation and create favorable conditions for convective precipitation. Radar and satellite imagery from 2018 to 2021 were collected and preprocessed for the development and evaluation of nowcasting models. 

\begin{figure*}[htbp]
    \centering
    \includegraphics[trim=50 100 250 50, clip,width=\textwidth]{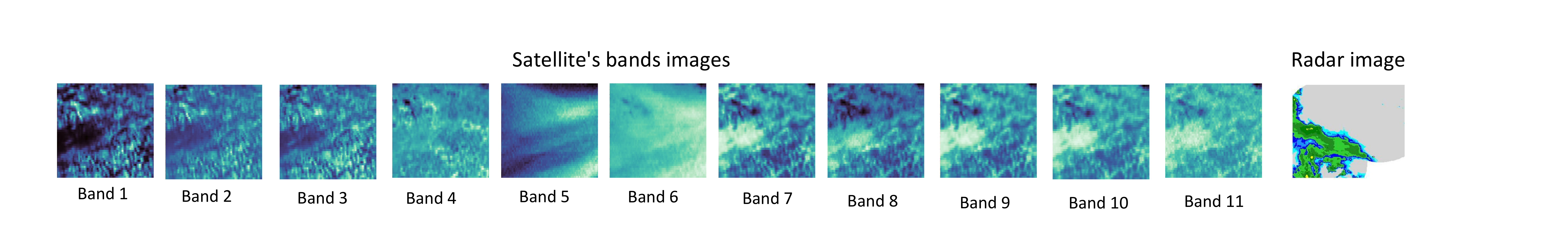}
    \caption{ Visualization of the 11 satellite bands and the radar image.}
    \label{fig:sat_radar_images}
\end{figure*}

\noindent\textbf{Radar Data:} High-resolution radar composites were provided by the German Meteorological Service (DWD) \cite{dwd_radar}, which were generated by merging data from multiple radar stations and the quality was enhanced by hydro \& meteo GmbH. The data include measurements from four radars: Neuheilenbach, Essen, Flechtdorf, and Hannover, with a spatial resolution of 1 km and a temporal resolution of 5 minutes. The dataset was cropped to cover roughly the NRW state, resulting in 288×288 pixel radar images, where each pixel corresponds to a precipitation value in mm/h, as shown in Fig.~\ref{fig:sat_radar_images}.

Following the classification defined by Barani Design Technologies \cite{barani2020rainrate}, rainfall intensities are divided into light (0–2.5 mm/h), moderate (2.5–7.5 mm/h), heavy (7.5–50 mm/h), and violent (50–200 mm/h) categories. Pixels with missing or undefined values were labeled as -999 and treated as a separate category.

\noindent\textbf{Satellite Data:} 
Satellite data were acquired from the SEVIRI Level 1.5 rapid scan product by EUMETSAT, offering 11 spectral bands with 5-minute temporal resolution \cite{eumetsat_seviri}. The original images, covering Europe and parts of the Middle East, were reprojected to EPSG:25832 (ETRS89 / UTM zone 32N) and cropped to the NRW state, yielding images of 47×92 pixels, as shown in Fig.~\ref{fig:sat_radar_images}.

The dataset includes spectral channels across the visible (VIS), near-infrared (NIR), infrared (IR), and water vapor (WV) ranges, supporting a wide range of meteorological observations. Note that VIS and NIR bands are only available during daytime.

\noindent\textbf{Band Histogram Analysis:} To explore potential correlations between satellite bands and rainfall intensity, histogram analyses were performed for light, moderate, heavy, and violent rain samples. For heavy rain events, two case studies on July 14, 2021 were analyzed using statistical tools including the Kolmogorov–Smirnov test and Kullback-Leibler divergence. The analysis revealed statistically significant differences between histograms of the two events across all bands, indicating that histogram-based features are not stable enough across samples for prediction. Consequently, raw satellite data were used as model input.

\subsection{Data Preprocessing and Normalization}

To ensure consistent and effective model training, several preprocessing and normalization steps were applied. First, radar images containing extreme outliers ($>$200 mm/h), accounting for 2.1\% of the data, were removed due to reliability concerns. To address class imbalance, only 20\% of no-rain images were retained. A 30-minute input window (6 consecutive images) was used, and sequences with missing radar or satellite files were discarded.

Radar values were normalized using a logarithmic transformation based on the precipitation maximum value (200): 

% \begin{algorithm}
% \caption{Normalization Procedure}
% \begin{algorithmic}[1]
% \State \textbf{Input:} Array of values \( x \)
% \State \textbf{Output:} Normalized array \( x_{\text{norm}} \)
% \For{each element \( x_i \) in \( x \)}
%     \If{\( x_i \geq 0 \)}
%         \State Shift value: \( x_i \gets x_i + 1 \)
%     \EndIf
%     \If{\( x_i = -999 \)}
%         \State Set value to zero: \( x_i \gets 0 \)
%     \EndIf
% \EndFor
% \For{each element \( x_i \) in \( x \)}
%     \State Normalize: 
%     \[
%     x_{\text{norm},i} = \log_{202}(x_i + 1)
%     \]
% \EndFor
% \State \Return \( x_{\text{norm}} \)
% \end{algorithmic}
% \end{algorithm}
\begin{equation}
x_{\text{norm}} =
\begin{cases}
\log_{202}(x + 2), & \text{if } x \geq 0 \\
0, & \text{if } x = -999
\end{cases}
\end{equation}

The base 202 accounts for the maximum value of 200, plus the shift (+2 total) to ensure numerical stability.

Satellite data, consisting of 11 bands, was normalized per band using Min-Max scaling:
\begin{equation}
X_{\text{norm}} = \frac{X - X_{\min}}{X_{\max} - X_{\min}}
\end{equation}
where $X_{\min}$ and $X_{\max}$ were determined from the training set for each band separately.

\subsection{Evaluation Metrics}
% To evaluate and compare the performance of the proposed models, two commonly used metrics in meteorological forecasting are employed: Critical Success Index (CSI), and Fractions Skill Score (FSS). These metrics capture different aspects of prediction quality and are well-suited to the specific challenges of precipitation nowcasting. Since predicting exact rainfall amounts (in mm/h) remains highly complex and is not the focus of this work, the evaluation is carried out using categorized precipitation intensities rather than continuous values.
To evaluate and compare the performance of the proposed models, two commonly used metrics in meteorological forecasting are employed: the Critical Success Index (CSI) and the Fractions Skill Score (FSS). These metrics capture different aspects of prediction quality and are well-suited to the specific challenges of precipitation nowcasting. The models are regression-based, outputting continuous rainfall estimates (mm/h). For evaluation purposes, these numerical outputs are discretized into precipitation intensity categories, on which the metrics are then computed.

\vspace{1ex}
\noindent\textbf{Critical Success Index (CSI):} The Critical Success Index (CSI), also known as the Threat Score, is a categorical metric widely used in weather prediction. It assesses the ability of the model to correctly forecast the occurrence of precipitation exceeding a predefined threshold. It is particularly relevant when precise rainfall magnitudes are less critical than the correct detection of heavy rain events.

The CSI is computed based on a binary classification framework using a contingency table:

\begin{equation} \text{CSI} = \frac{\text{TP}}{\text{TP} + \text{FP} + \text{FN}} \label{eq:csi} \end{equation}

Where TP (True Positives) refers to cases where the model correctly predicted the rain category (e.g., heavy or violent rain), FP (False Positives) indicates cases where the model predicted a rain category that did not actually occur, and FN (False Negatives) denotes cases where a rain category occurred but was not predicted by the model.
The CSI ranges from 0 to 1, where 0 indicates no skill and 1 represents a perfect score.

\vspace{1ex}
\noindent\textbf{Fractions Skill Score (FSS):} The Fractions Skill Score (FSS)\cite{fss} is a neighborhood-based metric that evaluates the spatial similarity between forecast and observed precipitation fields. Unlike pixel-wise scores, FSS accounts for small spatial shifts, making it especially suitable for evaluating precipitation forecasts where minor misalignments are common.

First, binary probabilities (BP) are computed using a rainfall threshold range [q1,q2):

\begin{equation}
\text{BP}_{i} =
\begin{cases} 
1 & \text{if } q_{1} \leq F_{i} < q_{2} \\
0 & \text{otherwise}
\end{cases}
\label{eq:BP}
\end{equation}

Where $F_{i}$  is the rainfall value at grid point $i$, and $q_{1}$, $q_{2}$ define the category bounds (in mm/h).

Neighborhood probabilities (NP) for predicted and observed fields are then calculated:

\begin{equation}
\text{NP}_i = \frac{1}{J} \sum_{j=1}^{J} \text{BP}_{ij}
\label{eq:NP}
\end{equation}

The Fractions Brier Score (FBS) is then calculated as:

\begin{equation}
\text{FBS} = \frac{1}{I} \sum_{i=1}^{I} \left( \text{NP}_{i,p} - \text{NP}_{i,o} \right)^2
\label{eq:FBS}
\end{equation}

The Worst Possible FBS (WFBS) is defined as:

\begin{equation}
\text{WFBS}= \frac{1}{I} \sum_{i=1}^{I} \left( \left(\text{NP}_{i,p}  \right)^2 + \left(  \text{NP}_{i,o}  \right)^2 \right)
\label{eq:WFBS}
\end{equation}

Finally, the FSS is computed as:

\begin{equation}
\text{FSS} = 1 - \frac{\text{FBS}}{\text{WFBS}}
\label{eq:FSS}
\end{equation}

Where I is the number of valid prediction-observation pairs, J the number of neighboring cells. ($NP_{i,p}$) and ($NP_{i,o}$) denote predicted and observed neighborhood probabilities, respectively.

The FSS ranges from 0 (no skill) to 1 (perfect skill). Values above 0.5 are typically interpreted as indicative of useful predictive skill. The FSS is particularly sensitive to neighborhood size: larger neighborhoods smooth out differences and often yield higher scores. In this work, a neighborhood size of 3 was selected to account for minor misalignments corresponding to ~1 km shifts for a single value in a radar prediction.

\subsection{Models}

Two models were developed based on U-Net architecture: a \textbf{radar-only model} and a \textbf{multimodal model} combining radar and satellite data. Each model was trained for three lead times (5, 15, and 30 minutes), resulting in six trained models.

\vspace{1ex}
\noindent\textbf{Radar-only Model:} This model uses a U-Net architecture similar to the RainNet model but uses 3D convolutions instead. The 3D convolutions on the radar input have a depth of 1. The input consists of six consecutive radar frames. 
For a 5-minute lead time, the model is given radar frames from \( t{-}30 \) to \( t{-}5 \) and predicts the radar frame at t (5 minutes after the last input). For a 15-minute lead time, the model takes \( t{-}40 \) to \( t{-}15 \) minutes of radar frames, and for a 30-minute lead time, it takes \( t{-}55 \) to \( t{-}30 \) minutes.

The encoder downsamples via 3D max pooling (2$\times$2$\times$1), while the decoder upsamples accordingly. Skip connections link corresponding levels. Each encoder level includes two Conv3D layers with ReLU activations, with the first layer doubling the channel size. In total, the model includes 20 Conv3D layers, four pooling layers, four upsampling layers, and four skip connections. Its output consists of continuous rainfall estimates in mm/h, which are subsequently mapped to precipitation categories for evaluation. Training was performed using the Adam optimizer (learning rate = 1e$^{-4}$) with scheduled decay at epochs 10, 30, and 40 over 50 epochs. The model has approximately 31.4 million parameters.

% \begin{figure*}[htbp]
%     \centering
%     \includegraphics[height=9cm, width=14cm]{chapters/pictures/Rainnet 3D radar2.png}
%     \caption{Modified RainNet for Radar Input}
%     \label{fig:rainnet_radar}
% \end{figure*}

\vspace{1ex}
\noindent\textbf{Multimodal Model:} The multimodal model has a similar architecture to the radar-only model, but takes a 12-channel input as depth: one radar band and 11 satellite bands per timestamp, and predicts the next radar frame at time \( t \). Satellite data was preprocessed (cropped to same radar area and upsampled with Lanczos interpolation) to match the radar resolution and area (288$\times$288). The resulting input volume spans six timestamps with 12 channels, yielding an input shape of 288$\times$288$\times$12. The radar-only model takes input of shape 6×288×288×1 (time × height × width × channels) while the multimodal takes 6×288×288×12.

\begin{figure*}[htbp]
    \centering
    \includegraphics[trim=750 250 250 100, clip,width=\textwidth]{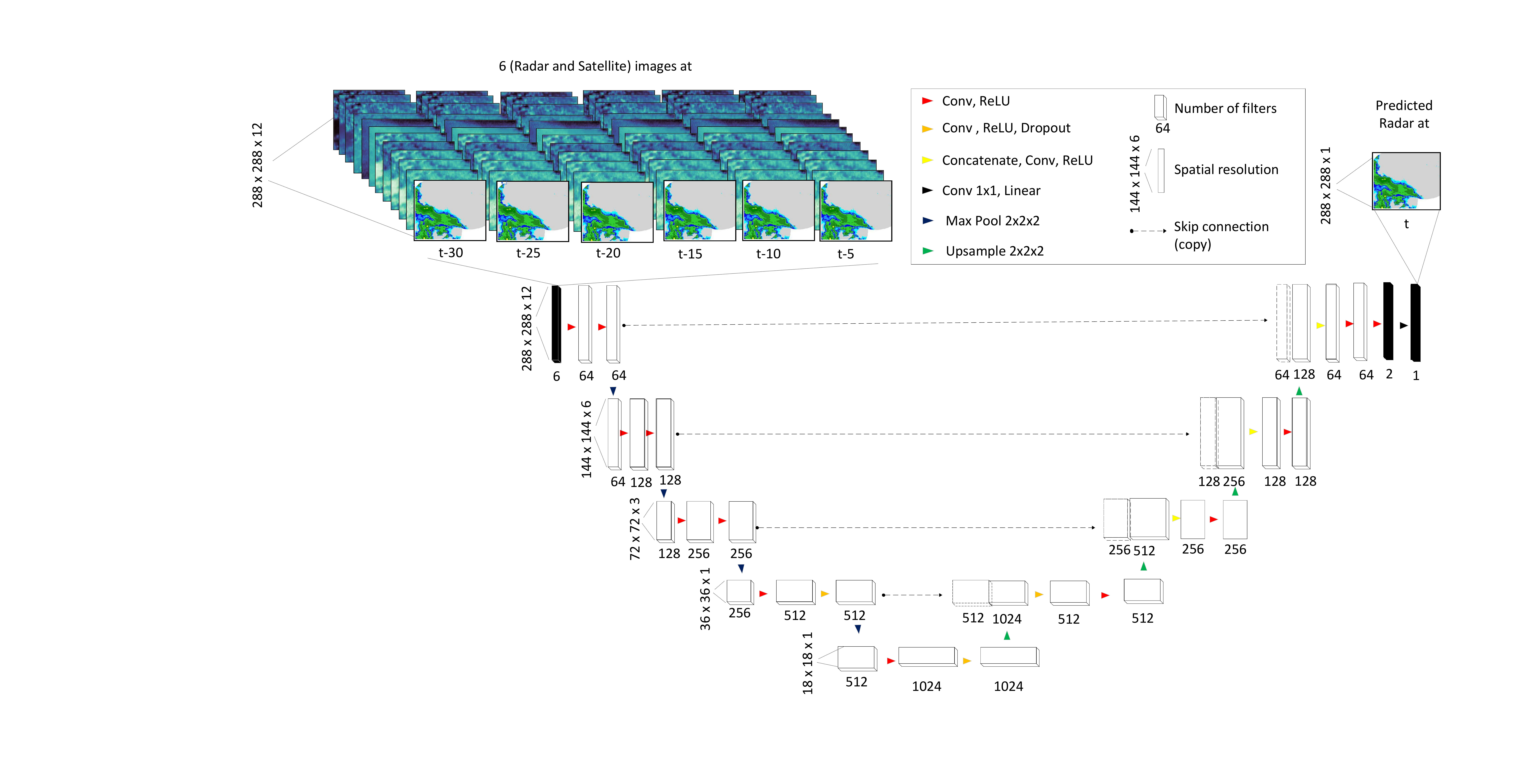}
    \caption{Architecture of the multimodal model. The model (for 5~min lead time) takes as input a sequence of six time steps (from $t{-}30$ to $t{-}5$), where each frame includes 12 channels: 1 radar image and 11 satellite bands. These inputs are processed by a 3D U-Net architecture consisting of 3D convolutional layers, max-pooling, and upsampling operations. The model outputs a single radar image predicting precipitation at $t$. Skip connections and concatenation operations are used between encoder and decoder stages to preserve spatial and temporal information.}
    \label{fig:rainnet_sat}
\end{figure*}

The temporal dimension (six frames) is halved in the encoder via pooling (2$\times$2$\times$2) and restored in the decoder through upsampling. The model outputs a single radar frame containing numerical precipitation estimates (mm/h). As shown in Fig.~\ref{fig:rainnet_sat}, the number of parameters increases to approximately 44 million due to the higher input dimensionality.

% \begin{figure*}[htbp]
%     \centering
%     \includegraphics[trim=60 250 250 190, clip,width=\textwidth]{chapters/pictures/Rainnet 3D Sat - paper.pdf}
%     \caption{Architecture of the multimodal model. The model (for 5~min lead time) takes as input a sequence of six time steps (from $t{-}30$ to $t{-}5$), where each frame includes 12 channels: 1 radar image and 11 satellite bands. These inputs are processed by a 3D U-Net architecture consisting of 3D convolutional layers, max-pooling, and upsampling operations. The model outputs a single radar image predicting precipitation at $t$. Skip connections and concatenation operations are used between encoder and decoder stages to preserve spatial and temporal information.}
%     \label{fig:rainnet_sat}
% \end{figure*}

\subsection{Loss Function}

% Two loss functions were evaluated: Mean Squared Error (MSE) and log-cosh. Based on experimental results, the log-cosh loss outperformed MSE in overall prediction quality.

% \begin{itemize}
%     \item \textbf{Mean Squared Error (MSE)}: Also known as L2 loss, this function penalizes larger errors more severely by squaring the difference between predictions and ground truth \cite{mse}. It is defined as:
%     \begin{equation}
%     \text{MSE}(y, \hat{y}) = \frac{1}{n} \sum_{i=1}^{n} (y_i - \hat{y}_i)^2
%     \label{eq:mse}
%     \end{equation}
%     where $n$ is the number of grid cells, $y_i$ is the true value, and $\hat{y}_i$ is the predicted value.

%     \item \textbf{Log-cosh Loss}: Adopted from the original RainNet model \cite{rainnet}, this function is less sensitive to outliers and behaves like MSE for small errors but more robust for larger ones. It is defined as:
%     \begin{equation}
%     \text{LogCosh}(y, \hat{y}) = \frac{1}{n} \sum_{i=1}^{n} \log\left(\cosh(\hat{y}_i - y_i)\right)
%     \label{eq:logcosh}
%     \end{equation}
%     with $\cosh(x) = \frac{e^x + e^{-x}}{2}$.
% \end{itemize}

Two loss functions were evaluated: Mean Squared Error (MSE) and log-cosh. Based on experimental results, the log-cosh loss outperformed MSE in overall prediction quality. The log-cosh loss was adopted from the RainNet model. log-cosh loss is less sensitive to outliers and behaves like MSE for small errors but more robust for larger ones. It is defined as:
    \begin{equation}
    \text{LogCosh}(y, \hat{y}) = \frac{1}{n} \sum_{i=1}^{n} \log\left(\cosh(\hat{y}_i - y_i)\right)
    \label{eq:logcosh}
    \end{equation}
    with $\cosh(x) = \frac{e^x + e^{-x}}{2}$.

\section{Experiments}

\subsection{Setup}
We trained and evaluated our models using radar and satellite data from North Rhine-Westphalia (NRW), Germany. Data were split as follows: summers of 2018 and 2019 for training, summer 2020 for validation, and summer 2021 for testing. A total of approximately 141,000 images were collected for the specified period, and these were used for training, evaluation, and testing. Training was conducted on NVIDIA A100 GPUs (80 GB), with the radar-only model using 2 GPUs and the multimodal model using 4 GPUs. Each epoch required 1 hour (radar-only) or 3.5 hours (multimodal). We evaluated models using Critical Success Index (CSI), and Fractions Skill Score (FSS).

% \subsection{Loss Function Comparison}
% To select the optimal loss function, we compared Mean Squared Error (MSE) and log-cosh using a fixed sample of 1000 sequences for 50 epochs. Log-cosh consistently outperformed MSE in both CSI and FSS metrics across heavy and violent rain categories (see Fig.~\ref{fig:mse_log_compare}), and was therefore used in all further experiments.

% \begin{figure}[htbp]
%     \centering
%     \includegraphics[width=1\linewidth]{chapters/pictures/results/MSE vs Log CSI.png}
%     \includegraphics[width=1\linewidth]{chapters/pictures/results/MSE vs Log FSS.png}
%     \caption{Comparison of CSI and FSS for MSE vs. log-cosh loss functions.}
%     \label{fig:mse_log_compare}
% \end{figure}

\subsection{Comparison Across Lead Times}
We compared the radar-only and multimodal models for 5, 15, and 30-minute lead times on the 2021 summer test set. Table~\ref{tab:leadtime_summary} summarizes CSI and FSS for heavy and violent rain.

\begin{table}[htbp]
\centering
\caption{Model performance across lead times on 2021 summer test set.}
\begin{tabular}{ccccc}
\toprule
\textbf{Lead Time} & \textbf{Category} & \textbf{Metric} & \makecell[c]{\textbf{Radar-only}\\\textbf{Model}} & \makecell[c]{\textbf{Multimodal}\\\textbf{Model}} \\
\midrule
\multirow{4}{*}{5 min} 
& \multirow{2}{*}{Heavy}   & CSI & 0.672 & \textbf{0.707} \\
&                          & FSS & 0.924 & \textbf{0.938} \\
\cmidrule(lr){2-5}
& \multirow{2}{*}{Violent} & CSI & 0.417 & \textbf{0.452} \\
&                          & FSS & 0.779 & \textbf{0.801} \\
\midrule
\multirow{4}{*}{15 min} 
& \multirow{2}{*}{Heavy}   & CSI & 0.368 & \textbf{0.379} \\
&                          & FSS & 0.677 & \textbf{0.689} \\
\cmidrule(lr){2-5}
& \multirow{2}{*}{Violent} & CSI & 0.030 & \textbf{0.039} \\
&                          & FSS & 0.104 & \textbf{0.129} \\
\midrule
\multirow{4}{*}{30 min} 
& \multirow{2}{*}{Heavy}   & CSI & 0.140 & \textbf{0.141} \\
&                          & FSS & \textbf{0.330} & 0.327 \\
\cmidrule(lr){2-5}
& \multirow{2}{*}{Violent} & CSI & 0.000 & \textbf{0.003} \\
&                          & FSS & 0.001 & \textbf{0.010} \\
\bottomrule
\end{tabular}
\label{tab:leadtime_summary}
\end{table}

At 5-minute lead time, the multimodal model improves heavy rain CSI by ~5\% (0.707 vs 0.672) and violent rain CSI by 0.035, showing better detection of intense cells. At 15 minutes, the gain drops to ~1\% (0.379 vs 0.368), and by 30 minutes, both models perform poorly, with only a marginal difference (0.141 vs 0.140).

As a result, the multimodal models outperformed the radar-only models in the lead times for both CSI and FSS evaluation metrics.

\subsection{Case Study: NRW Flood 2021}
To assess real-world performance, we evaluated the models on the flooding event of July 14, 2021, at 12:55. This event resulted in severe damage and loss of life across western Germany, underscoring the importance of accurate and timely rainfall predictions. The chosen radar image for this event contained significant heavy and violent rainfall, making it a valuable test case.

Table~\ref{tab:flood_case_combined} shows the evaluation of both models for all lead times. The multimodal model consistently outperformed the radar-only model across all metrics. Notably, at the 5-minute lead in the violent rain category, CSI jumps from 0.220 with radar-only to 0.384 with the multimodal model. The higher FSS (0.754 vs 0.526) further confirms the multimodal model’s superior spatial accuracy in pinpointing high-intensity rain regions.

\begin{figure*}[htbp]
    \centering

    \textbf{Radar-only Model Prediction (30 min lead time)}\\
    \includegraphics[width=0.98\textwidth]{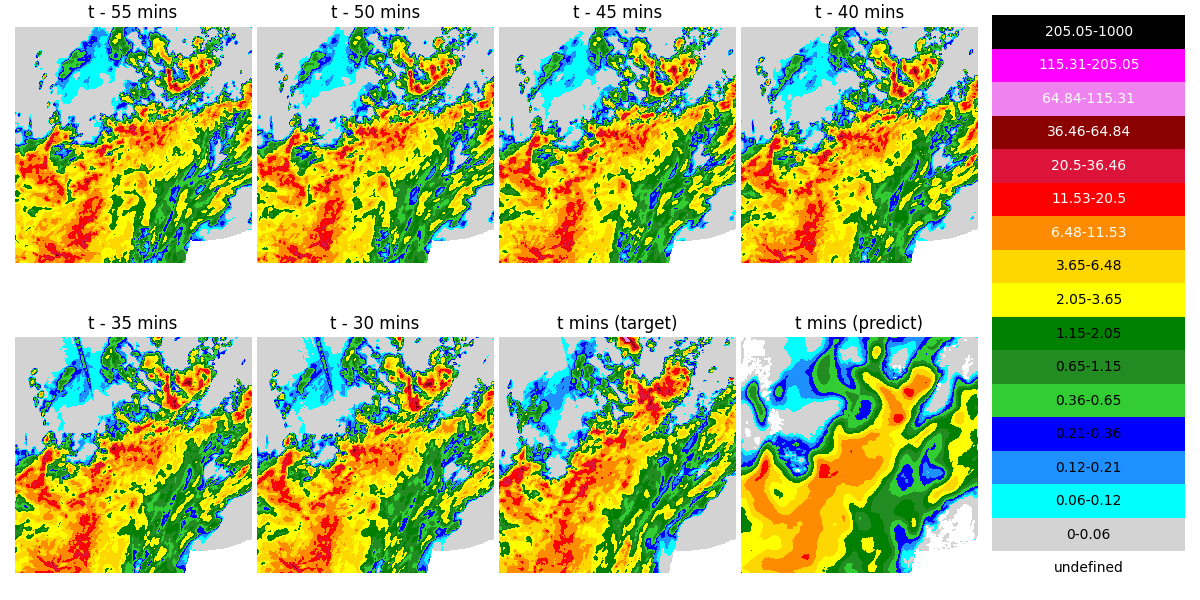}
    
    \vspace{1em}
    
    \textbf{Multimodal Model Prediction (30 min lead time)}\\
    \includegraphics[width=0.98\textwidth]{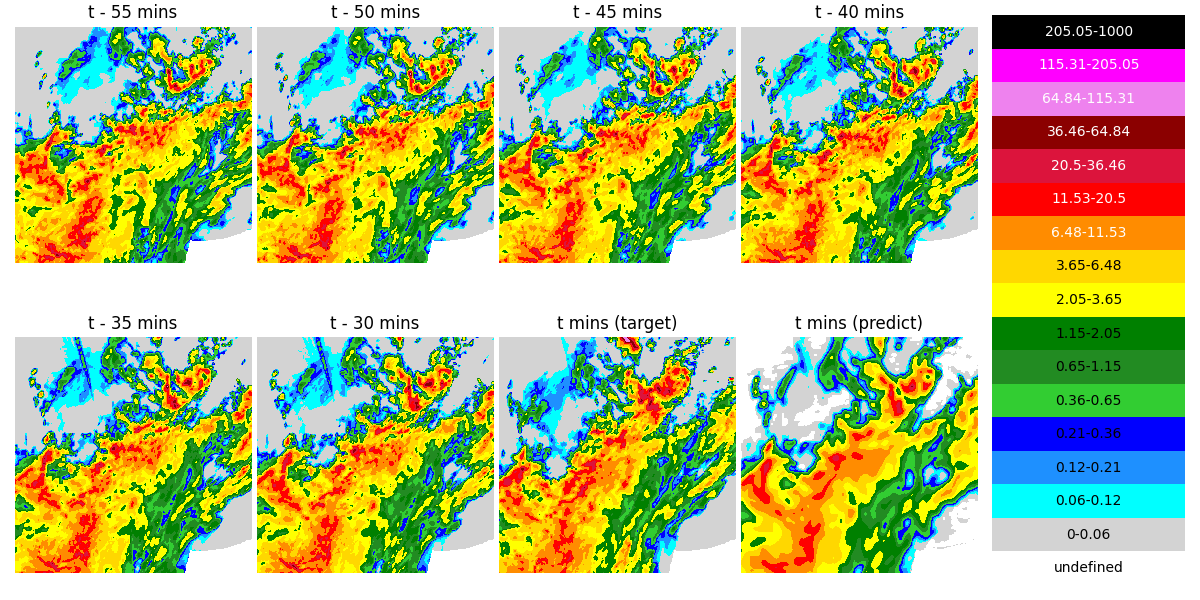}
    
    \caption{This figure shows the ground truth radar image (target) of a flooding event in North Rhine-Westphalia on July 14, 2021, along with the predictions at time~$t$ from two models: a Radar-only model and a Multimodal model. Both models use the same input sequence of radar images from $t{-}55$ to $t{-}30$ minutes. Additionally, the Multimodal model uses satellite images from the same time period. The color bar on the right indicates precipitation intensity in mm/h. The Multimodal model clearly outperforms the Radar-only model, particularly in regions of heavy rainfall, where the Radar-only model underestimates the intensity of the event.}
    \label{fig:flood_viz}
\end{figure*}

Fig.~\ref{fig:flood_viz} offers additional qualitative insight into the NRW flooding event. Even though Table~\ref{tab:flood_case_combined}  indicates only a marginal improvement for the multimodal model at the 30-minute lead time, Fig.~\ref{fig:flood_viz} shows that its prediction is qualitatively much closer to the ground truth than the radar-only model’s. The multimodal model effectively captured the shape, intensity, and distribution of heavy rain cells, which are crucial for issuing early warnings in operational settings. These improvements in qualitative accuracy were especially evident in urban areas that experienced flash flooding.

This case study highlights the added value of incorporating satellite imagery into short-term precipitation nowcasting, particularly for extreme weather events. The inclusion of atmospheric parameters from satellite data helped the model anticipate the buildup of convective storms, offering a more comprehensive understanding of evolving weather dynamics.

% In terms of qualitative analysis, predictions of the multimodal model more closely resembled the actual radar image. It effectively captured the shape, intensity, and distribution of heavy rain cells, which are crucial for issuing early warnings in operational settings. These improvements in qualitative accuracy were especially evident in urban areas that experienced flash flooding.

% To further evaluate robustness, we tested both models at longer lead times of 15 and 30 minutes on the same flood event. As expected, performance decreased at longer horizons, but the multimodal model continued to outperform the radar-only model, particularly for violent rain. Table~\ref{tab:flood_case_combined} summarizes the results. At 15 minutes, the multimodal model offered improved CSI and FSS for both heavy and violent rain. At 30 minutes, although both models failed to capture violent rain, the multimodal model provided slightly better localization of heavy precipitation zones. As shown in Fig. \ref{fig:flood_viz} the multimodal model shows slightly better results in the heavy rain category
% and improved qualitative visualization.

\begin{table}[htbp]
\centering
\caption{Model performance comparison for NRW flood event (2021-07-14, 12:55) across lead times.}
\begin{tabular}{ccccc}
\toprule
\textbf{Lead Time} & \textbf{Category} & \textbf{Metric} & \makecell[c]{\textbf{Radar-only}\\\textbf{Model}} & \makecell[c]{\textbf{Multimodal}\\\textbf{Model}} \\
\midrule
\multirow{4}{*}{5 min} 
& \multirow{2}{*}{Heavy}   & CSI & 0.801 & \textbf{0.822} \\
&                          & FSS & 0.962 & \textbf{0.970} \\
\cmidrule(lr){2-5}
& \multirow{2}{*}{Violent} & CSI & 0.220 & \textbf{0.384} \\
&                          & FSS & 0.526 & \textbf{0.754} \\
\midrule
\multirow{4}{*}{15 min} 
& \multirow{2}{*}{Heavy}   & CSI & 0.631 & \textbf{0.644} \\
&                          & FSS & 0.858 & \textbf{0.869} \\
\cmidrule(lr){2-5}
& \multirow{2}{*}{Violent} & CSI & 0.000 & 0.000 \\
&                          & FSS & 0.000 & 0.000 \\
\midrule
\multirow{4}{*}{30 min} 
& \multirow{2}{*}{Heavy}   & CSI & 0.414 & \textbf{0.419} \\
&                          & FSS & 0.662 & \textbf{0.664} \\
\cmidrule(lr){2-5}
& \multirow{2}{*}{Violent} & CSI & 0.000 & 0.000 \\
&                          & FSS & 0.000 & 0.000 \\
\bottomrule
\end{tabular}
\label{tab:flood_case_combined}
\end{table}

% \subsection{Effect of Neighborhood Size on FSS}

% We analyzed the effect of neighborhood size (1, 3, 5, 10 km) on FSS for the July 14, 2021 event. Fig.~\ref{fig:FSS_heatmap_summary} shows that larger neighborhoods increase FSS but reduce spatial precision. A size of 3 km was selected for all evaluations.

% \begin{figure}[htbp]
%     \centering
%     \includegraphics[width=0.8\linewidth]{chapters/pictures/results/FSS heatmap 5 min radar.png}
%     \caption{Effect of neighborhood size on FSS (5 min, radar-only model).}
%     \label{fig:FSS_heatmap_summary}
% \end{figure}

\section{Conclusion}
This work introduced a multimodal model combining radar and satellite data to improve short-term precipitation nowcasting at lead times of 5, 15, and 30 minutes. The models were trained on summer data from 2018–19, validated in 2020, and tested in 2021, focusing on convective rainfall.

The results show that the integration of satellite data with radar observations significantly enhances the capacity to detect and predict convective storms, particularly during their formative stages—where radar alone often falls short with longer lead times. Satellite imagery improves the model’s understanding of atmospheric conditions, leading to better forecasts of precipitation spatial distribution and intensity. Qualitative evaluations confirm these improvements, particularly during the July 2021 flood event as shown in Fig.~\ref{fig:compare_30_min} and Fig.~\ref{fig:flood_viz}.

%Unlike WeatherFusionNet \cite{weatherfusionnet}, which uses a physics-informed module, our approach demonstrates that a straightforward 3D-CNN fusion can already lead to significant improvements in the accuracy of heavy rainfall prediction.

%This work focused on comparing predictive performance across different data sources for a fixed architecture, previous studies have shown that RainNet outperforms ConvLSTM in similar radar datasets. Future work could include conducting such a comparison, expanding the training dataset, and exploring transformer-based architectures to further enhance nowcasting performance. Furthermore, investigating the individual contribution of satellite bands to prediction accuracy represents a promising direction for future research, especially as more computational resources become available.

% more data
% transformer
% which bad

This work focused on comparing predictive performance across different data sources using a fixed architecture. Future research could explore expanding the training dataset to include additional regions and employing more advanced architectures, such as transformers, to further improve nowcasting performance. Additionally, analyzing the individual contribution of satellite bands to prediction accuracy presents a promising direction.

\section*{Acknowledgment}
This paper, as part of the "heavyRAIN" project (funded by the mFUND program financed by the German Federal Ministry of Transport), explores satellite data collection, data preprocessing, and the development of multimodal models integrating radar and satellite data to enhance nowcasting accuracy and early warning system.

\bibliographystyle{IEEEtran}
\bibliography{IEEEabrv,./references.bib}

\end{document}